%% file: emnlp2022.tex
\pdfoutput=1
\documentclass[11pt]{article}

\usepackage[]{EMNLP2022}

\usepackage{times}
\usepackage{color}
\usepackage{latexsym}

\usepackage[T1]{fontenc}
\usepackage[utf8]{inputenc}

\usepackage{microtype}

\usepackage{inconsolata}

\usepackage{booktabs}
\usepackage{multicol}
\usepackage{graphicx}
\usepackage{amsmath}
\usepackage{tikz}
\usepackage{pgfplots}
\usepackage[capitalize,noabbrev]{cleveref}
\usepackage{tablefootnote}

\newcommand{\bs}[1]{\boldsymbol{#1}}
\newcommand{\func}[1]{\mathrm{#1}}
\newcommand{\mat}[1]{\mathbf{#1}}
\title{FiE: Building a Global Probability Space by Leveraging Early Fusion in Encoder for Open-Domain Question Answering}

\author{Akhil Kedia \and Mohd Abbas Zaidi \and Haejun Lee \\
        Samsung Research, Seoul \\
  \texttt{\{akhil.kedia, abbas.zaidi, haejun82.lee\}@samsung.com} \\}

\begin{document}
\maketitle

% Changes after Rebuttal
% Updated + in row iv) of Figure 3
% Added numbers to Table 4
% Correct y-axis label of Figure 3

\input{sections/abstract}
\input{sections/introduction}
\input{sections/proposed_method}
\input{sections/experimental_setup}
\input{sections/results}

\input{sections/ablation_studies}

\input{sections/analysis}
\input{sections/related}
\input{sections/conclusion}
\input{sections/limitations}

\bibliography{emnlp2022}
\bibliographystyle{acl_natbib}

\input{sections/appendix}
\end{document}

%% file: sections/abstract.tex
\begin{abstract}
Generative models have recently started to outperform extractive models in Open Domain Question Answering, largely by leveraging their decoder to attend over multiple encoded passages and combining their information. However, generative models tend to be larger than extractive models due to the need for a decoder, run slower during inference due to auto-regressive decoder beam search, and their generated output often suffers from hallucinations. We propose to extend transformer encoders with the ability to fuse information from multiple passages, using global representation to provide cross-sample attention over all tokens across samples. Furthermore, we propose an alternative answer span probability calculation to better aggregate answer scores in the global space of all samples. Using our proposed method, we outperform the current state-of-the-art method by $2.5$ Exact Match score on the Natural Question dataset while using only $25\%$ of parameters and $35\%$ of the latency during inference, and $4.4$ Exact Match on WebQuestions dataset. When coupled with synthetic data augmentation, we outperform larger models on the TriviaQA dataset as well. The latency and parameter savings of our method make it particularly attractive for open-domain question answering, as these models are often compute-intensive.
\end{abstract}
%  The latency and parameter savings of our method make it particularly useful in the context of deploying large language models to user devices, such as for question answering, dialog state tracking, dialog slot resolution, and dialog goal prediction

%% file: sections/introduction.tex
\section{Introduction}

Open-Domain Question-Answering is the task of answering an input question given a large external knowledge-base, such as the entire Wikipedia. 
%While very large LMs may directly produce an answer to the input query by utilizing the knowledge implicitly saved in their parameters\cite{gpt3, t5}, there are capacity constraints for the stored knowledge given the number of model parameters. Furthermore, given their large size, these models are also compute intensive to run.
This problem is typically approached by leveraging a retriever model to first retrieve a set of relevant documents/passages using some IR method, which are then passed on to a reader model \cite{orqa, dpr, realm, rag, ance, realmplusplus}.

The reader model then encodes all the passages through a transformer encoder separately, as transformers have quadratic computation with input sequence length. Extractive span-based methods \cite{orqa, dpr, realm, rag, ance} use these encoded representations for a per-passage selection of the answer span. Generative models such as FiD\cite{fid}, and derivative models utilizing a similar reader, such as \citet{fidkd, e2nr, emdr2, yono} use their decoder to attend to all the passages simultaneously by concatenating their encoded representations. The encoder-decoder cross attention enables the generative readers to fuse information globally across different passages.

So far, the existing works have tried to perform global information fusion by adding a decoder and hence adopting the generative approach. We hypothesize that this is not efficient for extractive tasks where the answer exists in the provided context passages. It may be helpful to restrict the answer probability space to the given context instead of using the universal vocabulary, which might also lead to issues such as hallucinations \cite{rel-hallu-6, rel-hallu-5, rel-hallu-4, rel-hallu-3}. Moreover, adding a decoder and performing auto-regressive answer generation increases the latency of the model. In this work, we propose to extend the transformer encoder of extractive models with the ability for early global fusion of information across input samples. We achieve this by adding global representation tokens to the input of the encoder, which can attend to all tokens in all passages with cross-sample attention. By doing so, not only do we remove the need for a decoder but also outperform existing generative approaches.
% , and hence reduce the parameter count by $2x$ and increase the inference throughput by avoiding auto-regressive generation.

\begin{figure*}[h!]
\includegraphics[width=\linewidth, trim={4.5cm 4.5cm 5.2cm 6.5cm}]{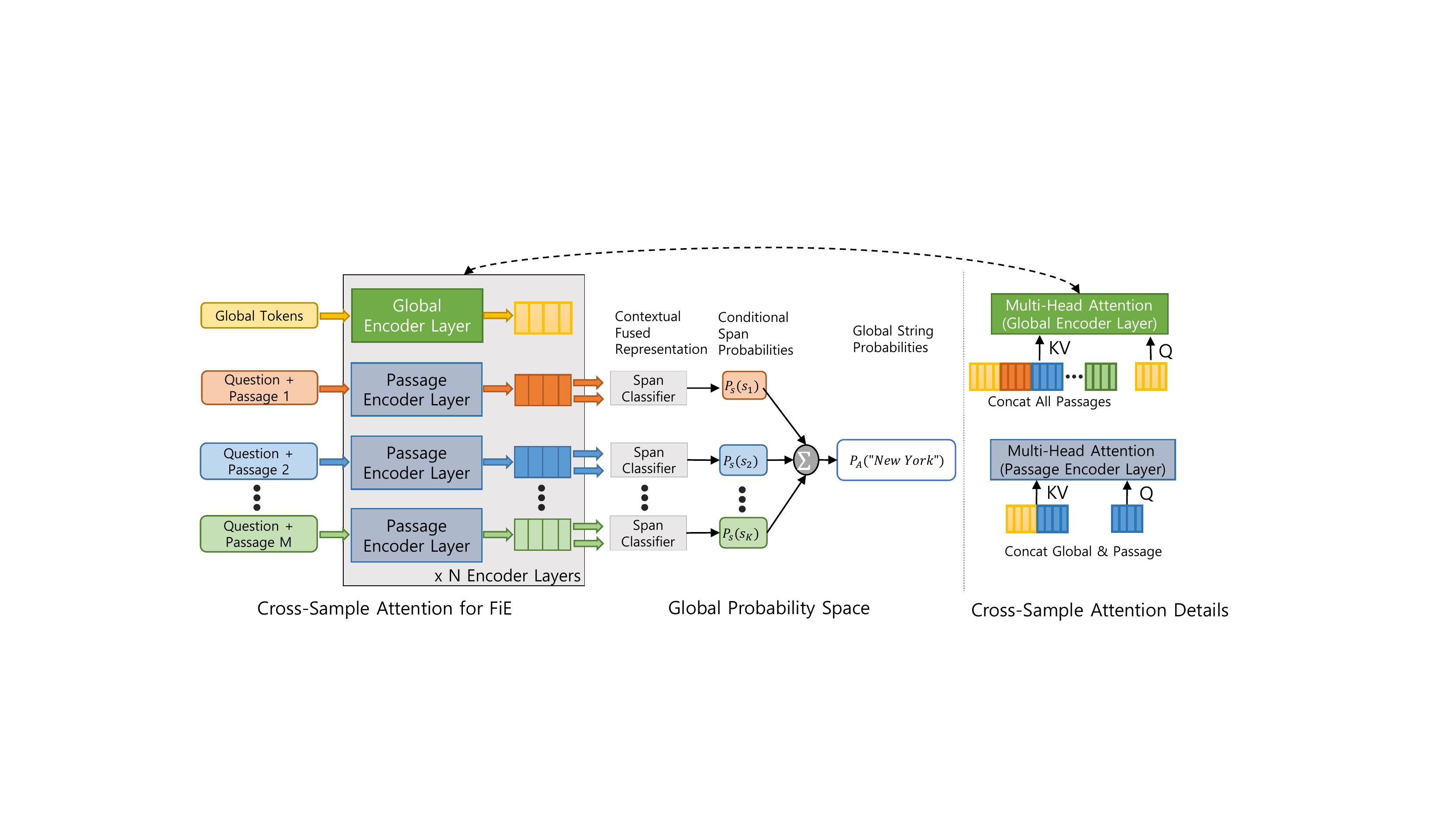} 
% \includegraphics[width=\linewidth, trim={4.5cm 6.2cm 5.2cm 6.5cm}]{fig/fie_diagram.pdf} 
% trim = {left,bottom,right,top}
\caption{The overall architecture of our proposed model FiE: Fusion in Encoder. The global representation tokens attend to all the tokens from all the passages, while all passages attend to themselves and to the global representation.}
\label{fie_architecture}
\end{figure*}

Extractive reader models typically \cite{orqa, dpr, realm} marginalize the probability of a given span on a per-passage level. However, several works \cite{prob-assum, cheng-etal-2021-unitedqa} show that more information may be gleamed from all the occurrences of an answer span across multiple passages. The authors change probability space from a per-passage level to a global level, achieving large performance gains. We also adopt a similar approach to take the best advantage of the global information flow in our model. Moreover, extractive models classify each token as the start/end of the answer span. The probability of a span is then the probability of its start token multiplied by its end token. This inherently assumes that the start and end probabilities of a span are independent of each other. We modify the probability space by directly calculating the score for each span rather than multiplying the start and end token scores. This approach enables us to calculate a separate score for each span and enables better aggregation of answer probability scores.

We evaluate our proposed method on the most commonly used open-domain datasets - Natural Questions \cite{nq}, TriviaQA \cite{joshi2017triviaqa} and WebQuestions \cite{webqa}. On Natural Questions, we outperform the previous best models having $4x$ more parameters by $2.5$ Exact Match, achieving a new state-of-the-art score of 58.4EM. We also achieve an improvement in score over the previous best models by $4.4$ Exact Match points on WebQuestions, and on TriviaQA outperform the best performing models when coupled with data augmentation. Our main contributions in this paper are -
\vspace{-1mm}
\begin{itemize}
    \item Fusing global information in the encoder by leveraging cross-sample attention, eliminating the need for a large and expensive decoder.
    \vspace{-1mm}
    \item Building a more robust global probability space for answer span probabilities.
\end{itemize}

%% file: sections/proposed_method.tex
\section{Proposed Method}
\label{proposed_method}

\subsection{Background Nomenclature}

Given input query tokens $\bs{q}$, $n$ contexts $\bs{C_j}$ with tokens $C_{i,j}$, and a transformer encoder $\func{E}$ with $l$ layers $\func{E_\ell}$, extractive methods pass them through the encoder to obtain encoded representations $\mat{h_{i,j}}$ for each token :
\begin{align}
\mat{h_{i,j}}=\func{E}(\bs{q,C_j})[i] \nonumber
\end{align}
Each token representation is then passed through a classifier $\func{W_{start}}$ to obtain the logit scores $s_{\func{start}}(i, j)$, for each token being the start of the answer span. These are then softmaxed across $i$ (on a per-passage level) to obtain probabilities, $p_\func{start}(i, j)$.
\begin{align}
    s_\func{start}(i, j) &= \func{W_{start}}(\mat{h_{i,j}}) \label{eq_old_st_class} \\ 
    p_\func{start}(i, j) &= \func{\sigma_j} (s_\func{start}(i, j)) \label{eq_old_st_prob},
\end{align}
where $\sigma_j$ is the softmax operation, across all tokens $i$ in the context $j$. Similarly, we obtain probabilities $p_\func{end}(i, j)$ for the end token. 
In \citet{prob-assum}, the authors instead softmaxed across all tokens $i$ across all contexts $j$, and showed that this global probability space helps.

The probability $p_\func{span}(st, en, j)$ for an answer span from $\bs{C_j}[st:en]$ are thereafter modeled as the product of the independent start and end token probabilities - 
\begin{align}
    p_\func{span}(st, en, j) = p_\func{start}(st, j)*p_\func{end}(en, j) \label{eq_old_prob}
\end{align}
% The highest scoring span is then selected as the answer span, and trained with cross-entropy loss.

Let $\bs{A}$ be the answer to the query $\bs{q}$, and let $Y_A$ be the set off all spans in all passages which exactly match the string $\bs{A}$ - 
\begin{align}
Y_A= \{ (st,en,j) \mid \bs{C_j}[st:en]=\bs{A}\} \nonumber
\end{align}

In \citet{prob-assum}, the authors show that aggregating the probabilities of all the spans that match the same string, $p_\func{string}(\bs{A})$, helps improve performance: 
\begin{align}
\begin{split}
p_\func{string}&(\bs{A}) = \\
\sum&\{p_\func{span}(st, en, j) \mid (st, en, j) \in Y_A \} \label{eq_prob_str}
\end{split}
\end{align}

\subsection{Cross-Sample Attention for Fusion in Encoder}
\label{sec:fie}
To extend the transformer with the ability for early global fusion of information in the encoder across input samples, we add $k$ extra ``global representation tokens'' $\bs{G}$ as an input to the encoder $\func{E}$. The input embeddings for these tokens are initialized with untrained extra vocabulary tokens, one for each global representation token. By modifying the transformer attention mechanism's key and value, these global representation tokens attend to all tokens in all passages with cross-sample attention, and all tokens can attend to these global tokens.

In the transformer attention block in each transformer layer $\func{E_l}$, let $\mat{Q_\ell}$, $\mat{K_\ell}$ and $\mat{V_\ell}$ be the attention's query, key and values respectively. Recall the attention function is then \cite{transformer}:
\begin{align}
\func{Attention}(\mat{Q_\ell},\mat{K_\ell},\mat{V_\ell})= \mathrm{softmax}(\frac{\mat{Q_\ell}\mat{K_\ell}^T}{\sqrt{d_K}})\mat{V_\ell} \nonumber
\end{align}

When the attention's query tokens $\mat{Q_\ell}$ are our global representation tokens, we change the key $\mat{K^G_\ell}$ and Value $\mat{V^G_\ell}$ to be the concatenation of all the encoded tokens of all the passages from the previous layer, as well as the global tokens, i.e., 
\begin{align}
\mat{K^G_\ell} = (\func{\oplus_i}(\func{E_{\ell-1}}(\bs{q}, \bs{C_j})))\oplus(\func{E_{\ell-1}}(\bs{G})) \nonumber
\end{align}
where $\oplus$ is the concatenation operation across all the passages and the global tokens. The same process is also repeated for the Value $\mat{V^G_\ell}$. This enables the representations for the tokens $\bs{G}$ to attend and fuse information from all the passages.

Similarly, tokens from any passage $\bs{C_j}$ can attend to the global tokens, along with other tokens from the same passage, i.e., the key for each passage is the concatenation of global tokens and passage tokens representations.
\begin{align}
\mat{K^i_\ell} = \func{E_{\ell-1}}(\bs{q}, \bs{C_j})\oplus \func{E_{\ell-1}}(\bs{G}), \nonumber
\end{align}

Because only the global representation tokens attend to all tokens, our method results in only $10\%$ overhead compared to a vanilla transformer, as we show theoretically and empirically in \cref{mem_lat_comp}.

\subsection{Global Probability Space}
\label{sec:glo_prob}

Our model now has information flow across samples, but when calculating the final probabilities in \cref{eq_old_st_class}, \cref{eq_old_st_prob} and \cref{eq_old_prob}, we ignore the information from presence of the answer span in other passages. \citet{prob-assum} attempts to fix this by modifying \cref{eq_old_st_prob} to softmax over all passages. However, due to \cref{eq_old_prob}, separate scores for start and end probability assigns incorrectly high score to long spans made by combining start/end of separate answer occurrence spans within the same passage.

To address this issue, we modify the span probability calculation to first concatenate the start and end embeddings, and classify this score for a span directly, and finally softmax across all the spans across all the passages - 
\begin{align}
\mat{h_{span}(st, en, j)} &= \mat{h_{st, j}} \oplus \mat{h_{en, j}} \nonumber \\ 
s_\func{span}(st, en, j) &= \func{W_{span}}(\mat{h_{span}(st,en,j)}) \nonumber \\ 
p_\func{span}(st, en, j) &= \func{\sigma_{st,en,j}} ( s_\func{span}(st, en, j)) \label{eq_new_span_prob}
\end{align}
where $\func{W_{span}}$ is non-linear classifier, $s_\func{span}(st, en, j)$ is the logit score for the span $\bs{C_j}[st:en]$, and the softmax is over all possible spans $st, en, j$ across all passages. 

In practice, because answer-span lengths are rarely longer than some constant $len_A$, we can assign a score of zero to all such spans in \cref{eq_new_span_prob} and skip calculating their actual scores and probabilities. 

Furthermore, following \citet{prob-assum}, we also aggregate the scores for all spans that are the same strings, as done in \cref{eq_prob_str}. This approach, combined with our changes to the probability space in \cref{eq_new_span_prob} and the global representation tokens introduced in \cref{sec:fie}, enables us to calculate probabilities for the answers in the global space of all strings in all passages. The span embeddings and corresponding probabilities are refined using information from other passages from the very first layer. The model is then trained with Maximum Marginal Likelihood (MML) objective of the correct answer string.

% Since we now have information flow across samples, after passing the span embeddings $s$ through the classifier, we now softmax the span logits over all the passages\cite{prob-assum} (rather than on a per-passages level), to obtain the probability of a span $P_s$. Furthermore, we aggregate the scores of all spans $s$ that are the same text string $A$, to obtain Global Probability of the string $A$ being the answer as $P_A=\sum{(P_s | \text{span } s \text{ is the String } A)}$. Our model is then trained with cross-entropy loss of the correct answer string.

%% file: sections/experimental_setup.tex
\begin{table*}[h!]
\begin{center}
\begin{tabular}{llclll}
\toprule
\textbf{Model} & \textbf{Model Type} &  \textbf{\# Params} & \textbf{NQ}  & \textbf{TQA}  & \textbf{WebQ}\\ 
\toprule 
\multicolumn{6}{l}{\textit{Base Models}} \\
REALM \cite{realm} & Extractive & 110M & 40.4  & - & 40.7 \\
DPR \cite{dpr} & Extractive & 110M & 41.5  &  56.8 & 34.6 \\
% E2NR \cite{e2nr} & Generative & 220M & 45.9 & 56.3 & - \\
ANCE \cite{ance} & Extractive & 110M & 46.0 & 57.5 & - \\
UnitedQA \cite{cheng-etal-2021-unitedqa} & Extractive & 110M & 47.7 & 66.3 & - \\
FiD \cite{fid} & Generative & 220M & 48.2 & 65.0 & 45.2  \\
KG-FiD \cite{kgfid} & Generative & 220M & 49.6 & 66.7 & - \\
FiD-KD \cite{fidkd} & Generative & 220M & 49.6 & 68.8 & 46.8 \\
EMDR$^2$ \cite{emdr2} & Generative & 220M & 52.5 & \textbf{71.4} & 48.7 \\
\midrule 
\textbf{FiE (Ours)} & Extractive & 110M & \textbf{54.9} & 68.2 & 50.8 \\
\textbf{FiE + PAQ (Ours)} & Extractive & 110M & 53.3 & 68.2 & \textbf{53.9} \\
% \midrule 
% \multicolumn{4}{l}{\textit{Larger Models}} \\
% \textbf{PAQ + FiE (Ours)} & Extractive & 110M & 52.4 & 68.2  & - \\
\midrule 
\multicolumn{6}{l}{\textit{Larger Models}} \\
RAG \cite{rag} & Generative & 400M & 44.5 & 56.1 & 45.2 \\
% E2NR \cite{e2nr} & Generative & 770M & 48.1 & 59.6 & - \\
R1-D1 \cite{r2d2} & Extractive & 330M & 50.8 & 65.0 & - \\
FiD \cite{fid} & Generative & 770M & 51.4 & 67.6 & - \\
UnitedQA \cite{cheng-etal-2021-unitedqa} & Extractive & 330M & 51.8 & 68.9 & 48.0
\\
KG-FiD \cite{kgfid} & Generative & 770M & 53.4 & 69.8 & - \\
FiD-KD \cite{fidkd} & Generative & 770M & 53.7 & 72.1 & - \\
UnitedQA \cite{cheng-etal-2021-unitedqa} & Hybrid & 1.87B& 54.7 & 70.5 & - \\
% UDT-QA \cite{udtqa} & Extractive & 330M & 55.2\textsuperscript{\textdagger} & - & 57.1\textsuperscript{\textdagger} \\
R2-D2 \cite{r2d2} & Hybrid & 1.29B& 55.9 & 69.9 & - \\
\midrule 
\textbf{FiE (Ours)} & Extractive & 330M & \textbf{58.4} & 71.6 & 52.4 \\
\textbf{FiE + PAQ (Ours)} & Extractive & 330M & \textbf{58.4} & \textbf{72.6} & \textbf{56.3} \\

% \textbf{PAQ + FiE (Ours)} & Extractive & 330M & 45.6 & - & - \\
\bottomrule
\end{tabular}
\end{center}
\caption{End-to-end Open QA Exact-Match results on Natural Questions (NQ), TriviaQA(TQA) and Web Questions(WebQ) test sets. PAQ is data-augmentation described in \cref{data_aug}. Our model outperforms all other models on Natural Questions and Web Questions, and all models of the same size on TriviaQA.}
\label{tab_main_result}
\end{table*}

\section{Experimental Setup}

\subsection{Datasets}

We perform our experiments on three of the most popular open domain question answering datasets - Natural Questions (NQ) \cite{nq}, TriviaQA \cite{joshi2017triviaqa} and WebQ \cite{webqa}. For NQ and TriviaQA, we used the short answer subsets processed by ORQA \cite{orqa}. For WebQ, we use the dataset version provided by EMDR2 \citet{emdr2}. We use the pre-processed Wikipedia dump from Dec. 20, 2018, provided by FiD-KD \citet{fidkd} as our external knowledge base. Dataset statistics and download links can be found in the supplementary material \cref{tab_dataset} and \cref{download_links}.

\subsection{Models}

We demonstrate the effectiveness of our method on a range of model sizes and capacities, from Bert \cite{devlin-etal-2019-bert} tiny (4M params), Electra \cite{clark2020electra} small (14M), base (110M) to large (330M). We use Electra model instead of BERT as \citet{clark2020electra} and UnitedQA \cite{udtqa} show that Electra outperforms BERT as a reader in open-domain QA, and previous SOTA R2-D2 \cite{r2d2} also used Electra. We carry out most of our ablations and analysis on Electra base, as it was faster to train compared to the large model, while still representing a reasonably strong backbone.

\subsection{Training Details}

We use all original hyper-parameters of Electra, use $k=10$ global tokens due to GPU memory constraints with the large model, and set the maximum number of answer tokens $len_A=15$. We use $n=100$ passages retrieved from the retriever of \citet{fidkd}, except for WebQuestions, for which we use the retriever from \citet{emdr2} with $n=50$. 

We run no hyper-parameter tuning and use all the original hyper-parameters except for the total number of training steps, details of which are in \cref{tab-hyperparam} in the supplementary. As our experiments are compute-intensive, we ran only a few run with varying seeds, as reported in \cref{tab_sup_std} in the supplementary. The experiments were run on 8x80GB Nvidia A100s with 800GB RAM and 4x32-core CPUs, and each experiment took around 1 day for NQ and 2 days for TriviaQA with large models. Inference was run on the same system, and took 2 minutes.

%% file: sections/results.tex
\section{Results}

\subsection{Open-domain QA}
\label{sec:res}

% \footnotetext{FiD results marked with \textsuperscript{*} are from \cite{emdr2}. UnitedQA results marked with \textsuperscript{+} are from \cite{udtqa}.}

As we show in \cref{tab_main_result}, both our base and large models outperform all previous approaches on the Natural Questions dataset, achieving a new State-of-the-art Exact Match scores of $58.4$ for the large model and $54.9$ for the base model, with gains of $2.5$ and $2.4$ EM over prior works. Our method even outperforms hybrid ensemble models \cite{cheng-etal-2021-unitedqa} with $6x$ more parameters while achieving much higher throughput during inference. 

We also outperform all previous models on WebQuestions, with scores $52.4$ and $50.8$ for large and base models, respectively, beating previous scores by $4.4$ and $2.1$ EM, respectively. On TriviaQA, our method outperforms all equal-sized models while being competitive with the current SOTA model \cite{fidkd}, with $2x$ more parameters, achieving a score of $71.6$ Exact Match with the large model.

We observed that generative models perform much better on TriviaQA. Let us compare the performance drop of extractive models on the NQ and TriviaQA datasets with respect to the generative Fid-KD model. In \cref{tab_main_result}, for each dataset where we have both NQ and TriviaQA results, we calculate the performance drop with respect to the generative FiD-KD model and take the average. We observe an average drop of 3.7 EM score on NQ compared to a much higher drop of 7.2 EM score on TriviaQA. This might explain the relatively smaller improvements of FiE for TriviaQA. 

% \begin{table}[h!]
% \begin{center}
% % \setlength{\tabcolsep}{0.3em}
% \begin{tabular}{llll}
% \toprule
% \textbf{Model}  & \textbf{NQ}  & \textbf{TQA}  & \textbf{WebQ}\\ 
% \toprule 
% % \multicolumn{4}{l}{\textit{Base Models}} \\
% \textbf{PAQ + FiE (Base)} & 52.4 & 68.2  & 53.9 \\
% % \midrule 
% % \multicolumn{4}{l}{\textit{Larger Models}} \\
% \textbf{PAQ + FiE (Large)} & 45.6 & 72.6 & 56.3 \\
% \bottomrule
% \end{tabular}
% \end{center}
% \caption{Effect of additional pretraining.}
% \label{tab_paq_result}
% \end{table}

\subsection{Data Augmentation for Open-domain QA}
\label{data_aug}

To study the impact of synthetic data augmentation on our model, we randomly sample 6M Question-and-Answer samples from PAQ \cite{paq}, use a pre-trained retriever from FiD-KD \cite{fidkd} to retrieve passages, and use this data to pre-train our model for 1 epoch. We show the results of this data augmentation in \cref{tab_main_result}.

We observe large gains of $3.9$ EM and $2.1$ EM in WebQuestions for large and base models, respectively, establishing new state-of-the-art results, perhaps due to the small size of the WebQuestions data training set. The results on TriviaQA are mixed - we observe no improvements for the base model, but the large model scores improve by $1$ EM, resulting in a new state-of-the-art score of $72.6$ EM. On NQ, we surprisingly observe a drop in model performance, perhaps because the model is over-fitting to the synthetic data.

% Most surprisingly, we observe large drops in model performance on NQ. From the much higher drop in performance for the large model of negative $12.8$ EM compared negative $1.6$ EM on base model, we conjecture that our model is over-fitting to the synthetic data, and is hence not able to learn as well on real NQ data, as the large model should be more prone to over-fitting.

\subsection{Open-domain QA across Model Sizes and Capacities}

Stronger models may not benefit as much from methods that improve the performance of weaker models. \cref{tab_abl_size} provides the performance of different extractive readers augmented with our global information fusion approach. These include the Bert \cite{devlin-etal-2019-bert} tiny, and Electra \cite{clark2020electra} small, base and large, ranging in size from 4M parameters to 330M. These results demonstrate the efficacy of our proposed approach across a wide range of different model sizes and capacities.

\begin{table}[h!]
\begin{center}
\begin{tabular}{llcc}
\toprule
\textbf{Model} & \textbf{Params} & \textbf{EM} & \textbf{$\Delta$ EM}\\ 
\toprule
Bert Tiny & 4M & 26.0 & +11.1 \\
% Bert Small
Electra Small & 14M & 43.1 & +2.6 \\
Electra Base & 110M & 54.9 & +6.9 \\
Electra Large & 330M & 58.4 & +7.7 \\
% Deberta Large
\bottomrule
\end{tabular}
\end{center}
\caption{Effect of Model Size on Performance (Exact Match) on Natural Questions test set. The last column is the improvement of our method over a baseline without any global representation tokens.}
\label{tab_abl_size}
\end{table}

%% file: sections/ablation_studies.tex
\section{Ablation Studies}

\subsection{Ablation of Model Components}

Both components of our approach, Cross-Sample Attention for Fusion in Encoder, and the Global Probability Space, provide significant improvements over the baseline (\cref{tab_ablation}). Our proposed probability space increases scores by $2.3$ EM on NQ, while fusion in encoder increases the scores by $3.2$ EM. 

Furthermore, our proposed approaches strongly complement each other. Global representation fusion allows information flow across documents to get global representations, and the global probability space provides a better training signal for FiE by necessitating information aggregation of the answer spans across documents. Adding both of these together results in a total increase of $9.2$ EM.

\begin{table}[h!]
\begin{center}
\begin{tabular}{lc}
\toprule
\textbf{Model} & \textbf{EM(NQ)}\\ 
\toprule
Baseline (Electra base) & 45.7 \\
Baseline + global prob. space & 48.0 \\
Baseline + global repr. fusion & 48.9 \\
Baseline + both (FiE) & \textbf{54.9} \\
\bottomrule
\end{tabular}
\end{center}
\caption{Ablation of FiE Model Components on Natural Questions with Electra Base model, Exact Match scores.}
\label{tab_ablation}
\end{table}

\subsection{Alternatives for Global Representation Fusion}
\begin{figure}[hbt]
\includegraphics[width=\linewidth]{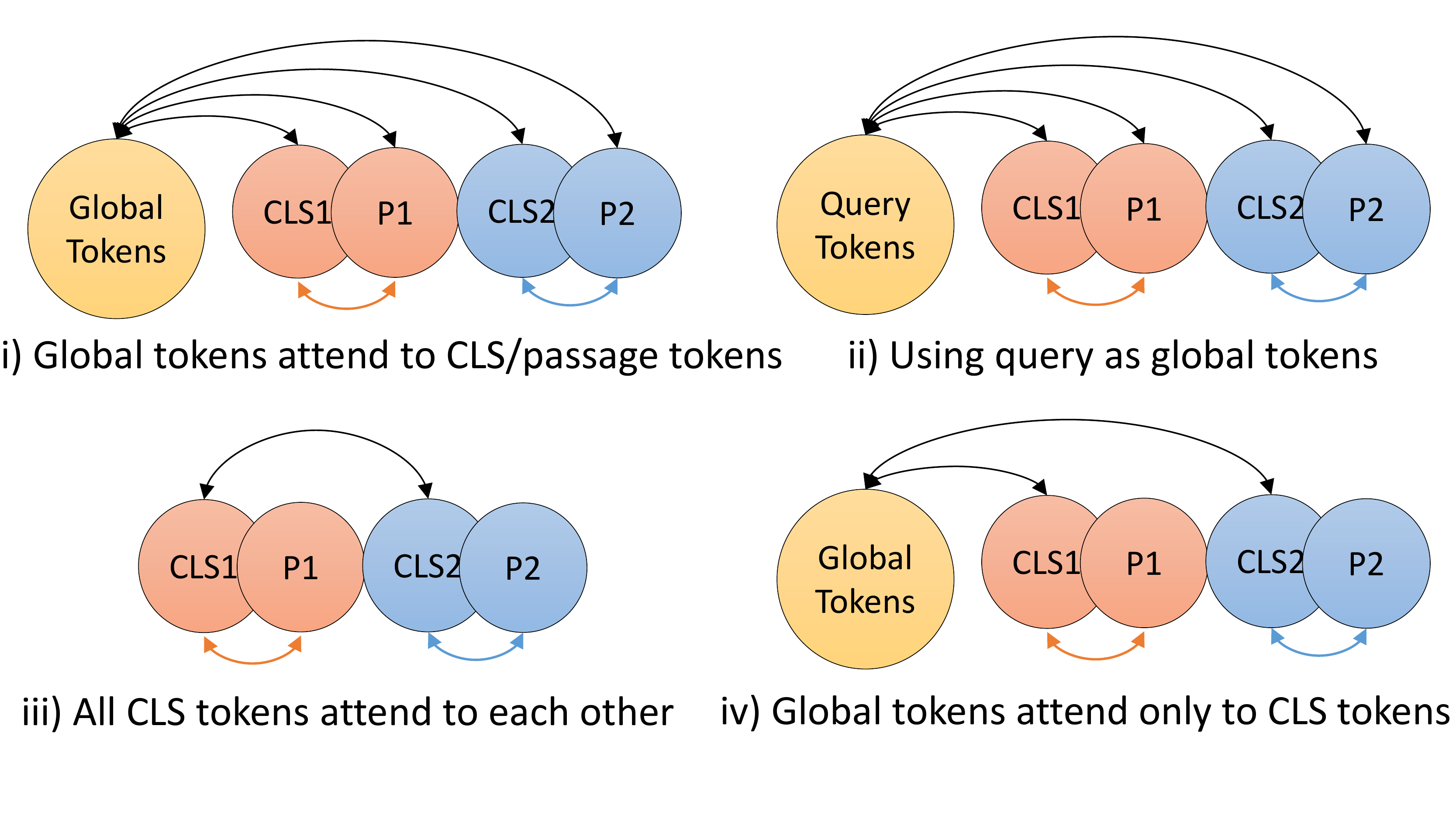} 
\caption{Alternatives for Global Representation Fusion}
\label{global_fusion}
\end{figure}

We use cross-sample attention to enable global information fusion in the encoder. In addition to the approach described in \cref{sec:fie}, we also study different alternative variants of the fusion mechanism, as shown in \cref{global_fusion}. Using the query token as input for the global tokens (instead of separate trained embeddings for each global token) may allow for better contextualization(\cref{global_fusion} (ii)). 

Alternatively, as the ``CLS'' token can capture some information for the entire passage, simply allowing all the ``CLS'' tokens to attend to one another can enable some global information flow(\cref{global_fusion} (iii)). Lastly, we restrict the cross attention of global tokens, such that they attend only to the CLS tokens of different passages(\cref{global_fusion} (iv)).

\begin{table}[h!]
\begin{center}
\begin{tabular}{lc}
\toprule
\textbf{Model} & \textbf{EM(NQ)}\\ 
\toprule 
i) Ours (FiE) & \textbf{54.9} \\
ii) Using query as global tokens & 52.2 \\
iii) All CLS attend to each other & 51.5 \\
iv) Global toks attend only to CLS & 50.0 \\
No fusion & 48.0 \\
Simple concatenation (10 context) & 41.6 \\
\bottomrule
\end{tabular}
\end{center}
\caption{Exact Match scores of alternatives to our encoder fusion tokens, with Electra Base model, on Natural Questions Test set. All models were trained with the global probability space. The numbers at the beginning of the rows correspond to the figures shown in \cref{global_fusion}.}
\label{tab_alternatives_fusion}
\end{table}

The results corresponding to these variations are reported in \cref{tab_alternatives_fusion}. Using dedicated tokens for the global tokens, rather than re-using the query tokens, results in better performance. We conjecture that it is easier for the model to learn how to fuse information if these tokens are consistently the same across different examples, rather than changing for each query. All these approaches improve the performance over the baseline model (no fusion), highlighting the importance of enabling information fusion in the encoder itself. Moreover, the model performance keeps increasing as we increase the scope of information fusion.

We also tested a simple baseline of concatenating the input passages. Electra is trained with maximum sequence length $512$. Since $10$ concatenated passages have a sequence length approximately $1500$, we extended and re-trained the position embeddings. This method obtained $41.6$ EM on NQ, as shown in \cref{tab_alternatives_fusion}, compared to our method's score of $49.1$ EM when given $10$ passages as shown in \cref{tab_abl_num_pas}. We conjecture that such a low score may be due to non-pretrained position embeddings, suggesting that approaches such as concatenating are perhaps best used with models trained with longer sequence length. Our method on the other hand, suffers from no such limitation.

\subsection{Alternatives for Global Probability Space}
\begin{figure}[hbt]
\includegraphics[width=\linewidth]{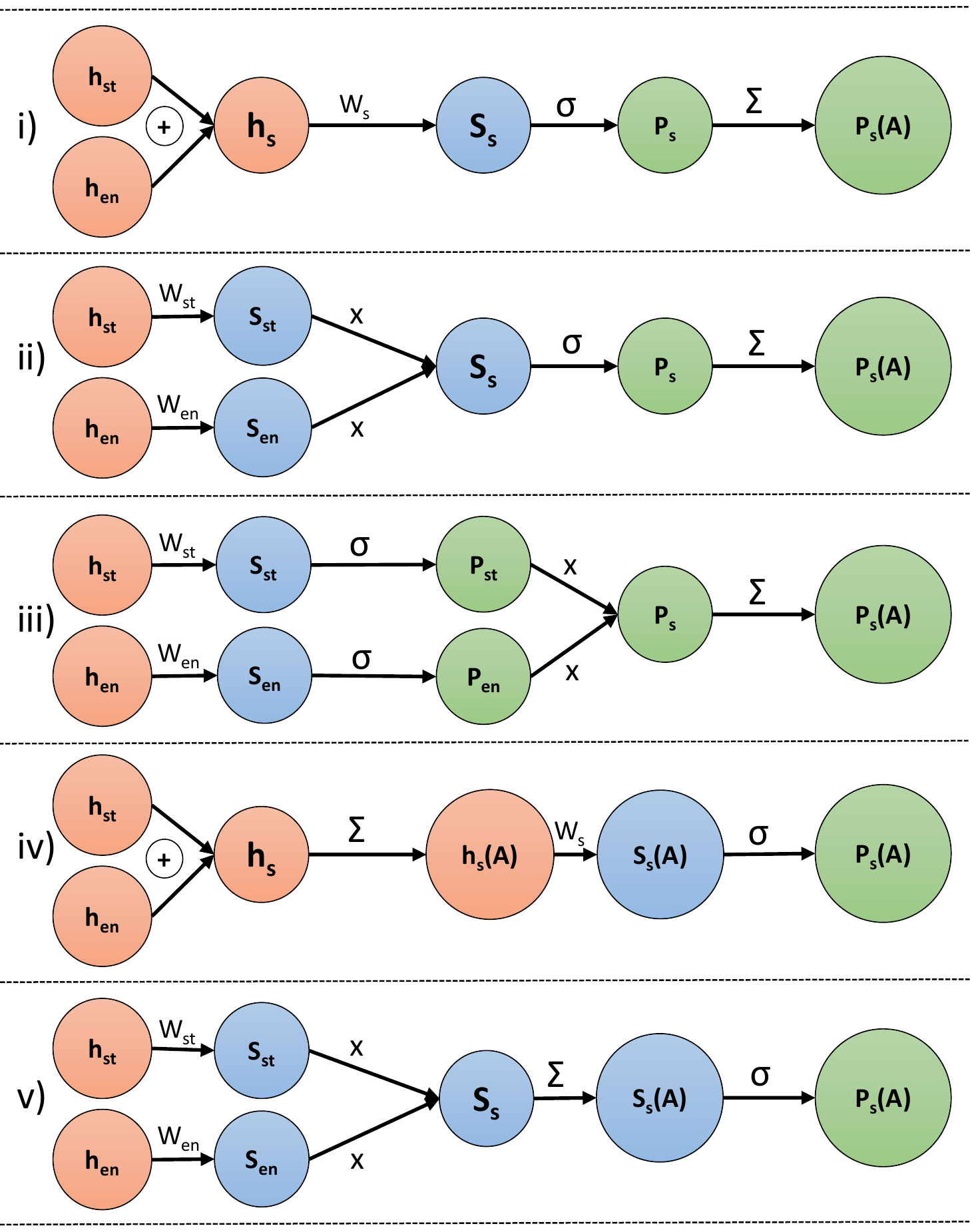} 
\caption{Alternatives for Global Probability Space. ($\oplus$: Concatenation, $\func{W}$: classifier, x: multiplication, $\sigma$: softmax,  $\sum$: Summation, $\mat{h}$: encoding, $s$: logit score (obtained using a classifier), $p$: probability (calculated via softmax)), $st$: start position, $en$: end position, $\bs{A}$: answer string)}
\label{global_prob}
\end{figure}

We investigate several alternatives for our global probability space for span score calculation (\cref{global_prob}). \cref{global_prob} (i) shows our approach, as described in \cref{sec:glo_prob}. In ``Non-conditional start \& end logits'' (\cref{global_prob} (ii)), we consider separately calculating the start and end span scores. In ``Separate start and end probability space'' (\cref{global_prob} (iii)), the start and end tokens are separately softmaxed across all possible tokens, as done in \citet{prob-assum}.

In ``string probability space'' (\cref{global_prob} (iv, v)), we make our probability space the space of all strings, softmaxing across the aggregated scores of all strings. We have two variants, in ``Span representation sum'' (\cref{global_prob} (iv)), the score for a string is classified by the aggregated embeddings of all spans. \cref{global_prob} (v) shows ``Logits sum'', where the string score is calculated by aggregating the scores of the spans.

Our probability space and score calculation approach outperforms all the other approaches (\cref{tab_alternatives_prob}). \citet{prob-assum}'s separate start and end probability space tends to incorrectly assign a high probability to spans made by combining start and end probabilities of different occurrences of the answer string. The string probability space methods also under-perform, perhaps because they aggregate before the softmax.
 
\begin{table}[h!]
\begin{center}
\begin{tabular}{lc}
\toprule
\textbf{Model} & \textbf{EM}\\ 
\toprule 
i) \hspace{2pt} Ours & \textbf{52.2} \\
ii) Non-conditional start \& end logits & 50.8 \\
iii) Separate start and end prob space & 48.1 \\
iv) Span repr. sum + string prob space  & 47.3 \\
v) \hspace{1pt} Logits sum + string prob space & 44.1 \\
\bottomrule
\end{tabular}
\end{center}
\caption{Exact Match scores of alternatives to the global probability space, with Electra Base model, on Natural Questions Test set. All models were trained with using query as global tokens.}
\label{tab_alternatives_prob}
\end{table}
 
\subsection{Combination of HardEM and Maximum Likelyhood Objectives}

\begin{table}[h!]
\begin{center}
\begin{tabular}{lc}
\toprule
\textbf{Model} & \textbf{EM(NQ)}\\ 
\toprule 
MML & \textbf{52.2} \\
HardEM max & 49.5 \\
HardEM min & 49.7 \\
HardEM 80\% probability mass & 49.9 \\
\bottomrule
\end{tabular}
\end{center}
\caption{Exact Match scores of alternatives to the MML training objective, with Electra Base model, on Natural Questions Test set. All models were trained with using query as global tokens.}
\label{tab_alternatives_obj}
\end{table}

Previous work \cite{prob-assum, cheng-etal-2021-unitedqa} has shown that adding global HardEM objective, using the maximum logit in positive paragraphs, may yield better results. We combine multiple variations of this HardEM with $0.1$ weightage added to our MML objective. All variations of HardEM decrease model performance (\cref{tab_alternatives_obj}). 

Because the HardEM objective may be more susceptible to false-positive occurrences in the paragraphs, it may negatively impact performance when fusing across passages. Vanilla MML assumes all spans are correct, while HardEM assumes only one is. In our modeling, the model is free to decide the relative correctness of each answer occurrence, with direct supervision only from the answer string without any additional assumptions.

%% file: sections/analysis.tex
\section{Analysis}

\subsection{Information Content of Global Tokens}

The global tokens were motivated to enable information flow across passages, by gathering information from all tokens. A very pertinent information to gather would possibly be the answer representation. To analyze the information content of global representation tokens, we rank all the tokens $C_{i,j}$ in the input passages by the similarity of their final encoded representations $\mat{h_{i,j}}$ to the final encoder representations of the global tokens $\mat{h_G}$. Let us call a token an ``answer token'' if it is present in the ground truth answer $\bs{A}$. 

We find that in $49\%$ of examples from NQ test set for Electra Base ($53\%$ for Large), the most similar input token to the global tokens is an answer token. This increases to $70\%$ if we only consider the examples the model answers correctly. Furthermore, for $43\%$ of examples, all the answer tokens are present in the 10 tokens most similar to global tokens after removing duplicates. Even without any explicit training objective to do so, the global tokens implicitly gather the answer representation.

\subsection{Importance of Global Tokens}

The model places a large importance on the global tokens, with the average attention to global tokens $2.4$ and $2.8$ times expected attention value, as shown in \cref{tab_glo_ana_attn} using NQ test set. This is even higher than the attention to the query tokens, and $17\%$ of the attention probability mass is on the global tokens, despite the number of the global tokens being only $6\%$ of the input sequence length.  

\begin{table}[h!]
\begin{center}
\begin{tabular}{lccc}
\toprule
\textbf{Model Size} & \textbf{Base} & \textbf{Large}\\ 
\toprule 
Attn. to query token & 2.1x & 2.2x \\
Attn. to global token & 2.4x & 2.8x \\
Attn. prob. mass query token & 14\% & 17\% \\
Attn. prob. mass global token & 17\% & 18\% \\
\bottomrule
\end{tabular}
\end{center}
\caption{Analysis of attention to global tokens. The first two rows show the ratio of attention to global/query tokens compared to expected average attention. The last two lines show the total attention probability mass.}
\label{tab_glo_ana_attn}
\end{table}

\subsection{Cross-sample Fusion by Global Tokens}

We use attention-rollout \cite{attn_flow} to measure how much cross-sample fusion is enabled by our model. Attention rollout models the information flow via attention as a directed graph through the layers, and assumes the representations of input tokens are linearly combined through the layers based on the attention weights. In our model, effective attention to other passages is achieved via passage tokens attending to the global tokens, which in turn attend to other passages. 

For the base model, we find that $46\%$ of the attention rollout probability mass for a given input passage is on other passages. This increases to $69\%$ for the large model, clearly demonstrating that our method successfully achieves cross-sample fusion.

\subsection{Effect of Number of Global Tokens}

\begin{figure}[h!]
\centering
\footnotesize
\begin{tikzpicture}
\begin{axis}[
    name=axis1,
    xlabel={Number of Fusion Tokens ->},
    ylabel={Exact Match (NQ) ->},
    height=5.5cm,
    xmin=-10, xmax=110,
    ymin=39.9, ymax=42.9,
    xtick={0,10,25,50,100},
    ytick={40,40.5,41,41.5,42,42.5,43},
    legend pos=south east,
    ymajorgrids=true,
    grid style=dashed,
    ylabel near ticks
]

\addplot[color=blue,mark=square,]
coordinates {(0,40.5)(10,41.6)(25,41.8)(50,42.3)(100,42.3)};
\end{axis}
\end{tikzpicture}

\caption{Effect of Number of Global Tokens on Model performance, in Exact-Match on NQ test set, with Electra-Small Model.}
\label{tab_abl_num_glo}
\end{figure}
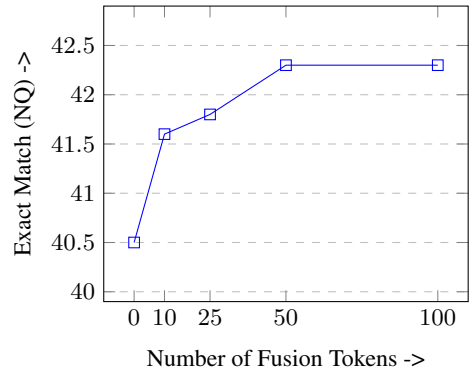
 
\cref{tab_abl_num_glo} provides variation in the model performance with the number of global tokens. More global tokens will lead to more model capacity for fusion across passages, and the results show that increasing the number of global representation tokens leads to better performance. Although we observe significant improvements up to 50 tokens, we use 10 global tokens in all other experiments due to GPU memory constraints with the large model. 

\subsection{Effect of Number of Passages}

% \begin{figure}[h!]
% \centering
% \footnotesize
% \begin{tikzpicture}
% \begin{axis}[
%     name=axis1,
%     xlabel={Number of Passages ->},
%     ylabel={Exact Match (NQ) ->},
%     height=6.0cm,
%     xmin=-5, xmax=110,
%     ymin=30, ymax=57,
%     xtick={1,5,10,20,50,100},
%     ytick={32,35,38,41,44,47,50,53},
%     legend pos=south east,
%     ymajorgrids=true,
%     grid style=dashed,
%     ylabel near ticks
% ]

% \addplot[color=blue,mark=square,]
% coordinates {(1,33.4)(5,47.6)(10,49.1)(20,51.3)(50,53.8)(100,54.9)};
% \addlegendentry{NQ}
% \addplot[color=red,mark=triangle,]
% coordinates {(1,38.2)(5,45.7)(10,47.1)(20,48.4)(50,50.8)};
% \addlegendentry{WebQ}
% \end{axis}
% \end{tikzpicture}
% \caption{Effect of Number of Passages, in Exact-Match on NQ and WebQ test set, with Electra-Base Model.}
% \label{tab_abl_num_pas}
% \end{figure}

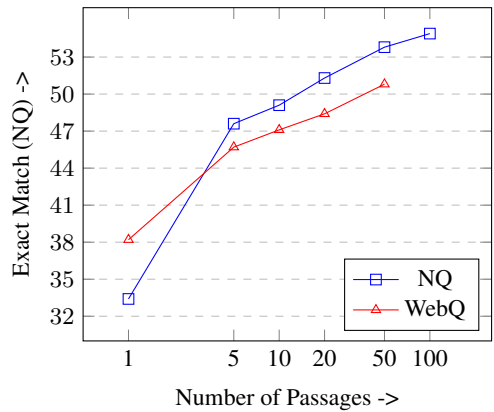
\begin{figure}[h!]
\centering
\footnotesize
\begin{tikzpicture}
\begin{axis}[
    name=axis1,
    xlabel={Number of Passages ->},
    ylabel={Exact Match (NQ) ->},
    height=6.0cm,
    xmin=-1, xmax=8,
    ymin=30, ymax=57,
    xtick={0,2.32,3.32,4.32,5.64,6.64},
    xticklabels={1,5,10,20,50,100},
    ytick={32,35,38,41,44,47,50,53},
    legend pos=south east,
    ymajorgrids=true,
    grid style=dashed,
    ylabel near ticks
]

\addplot[color=blue,mark=square,]
coordinates {(0,33.4)(2.32,47.6)(3.32,49.1)(4.32,51.3)(5.64,53.8)(6.64,54.9)};
\addlegendentry{NQ}
\addplot[color=red,mark=triangle,]
coordinates {(0,38.2)(2.32,45.7)(3.32,47.1)(4.32,48.4)(5.64,50.8)};
\addlegendentry{WebQ}
\end{axis}
\end{tikzpicture}
\caption{Effect of Number of Passages on a logarithmic scale, in Exact-Match on NQ and WebQ test set, with Electra-Base Model.}
\label{tab_abl_num_pas}
\end{figure}

Increasing the number of passages will directly provide more information to the model, allowing potentially better results. However, this may come with increased false positives and noise from irrelevant passages. \citet{wang-etal-2019-multi} shows that the performance of extractive models peaks around 10 to 20 passages. 

\cref{tab_abl_num_pas} shows that the performance of our proposed model consistently improves with increase in the number of passages, across multiple datasets. With the addition of global information fusion, our extractive models are able to utilize the extra information from multiple passages and hence improve the overall performance. The performance does not seem to have saturated, and seems to almost be in a power-law relationship to the number of passages. We only use 100 passages in all our experiments for fair comparison with prior works (50 for WebQ).
 

%% file: sections/related.tex
\section{Related Works}
\subsection{Open-domain Reader Models}
Reader models for Open-domain QA are required to read multiple documents (often more than 100 passages) \cite{fid,r2d2} to avoid missing the target passage from the large-scale knowledge base. Reading passages jointly at such a scale using a transformer encoder is computationally intractable due to the quadratic complexity of the transformer. To reduce the complexity, Knowledge-GPT \cite{knowledgegpt} selects key sentences from multiple passages and then encodes them all joint using GPT \cite{radford2019language}. OrQA \cite{orqa}, REALM \cite{realm}, and RAG \cite{rag} encode each passage separately and marginalize the predicted answer probabilities. However, these early approaches perform poorly due to a lack of information exchange across multiple passages for more contrastive representations during a prediction.

\subsection{Information Fusion for Reader Models}

Cross-sample information fusion in reader models is most commonly achieved by letting the decoder attend to the encoded inputs of multiple passages, as first proposed by FiD \cite{fid}, and then adopted by \citet{fidkd, emdr2, kgfid, r2d2, cheng-etal-2021-unitedqa}. Our proposed model achieves early fusion by adding global tokens to the encoder, ensuring the encodings of different sample passages are aware of global information.

BigBird \cite{bigbird}, Longformer \cite{longformer}, LongT5 \cite{longt5} focused on efficient attention to long sequences by allowing only a few tokens to have the full attention on all tokens in a single input passage, but these approaches did not consider fusing information across samples and often require expensive pre-training. 

A contemporaneous work Zcode++ \cite{zcode_plus_plus} proposes encoder fusion by concatenating all input tokens in the last encoder layers, also called fusion-in-encoder(FiE). However, the quadratic complexity of attention results in an overhead of $8x$ compared to our method, as shown in \cref{mem_lat_comp}.

\subsection{Global Probability Space for Readers}

To deal with multiple input passages, initial approaches for extractive models used a pipeline first to select a paragraph and then extract the answer \cite{realm, dpr, ance}. \cite{clark-gardner-2018-simple, wang-etal-2019-multi} added global normalization where the answer probability was normalized over multiple candidates, while \cite{wang2018evidence} also aggregated answer confidence from multiple passages based on strength and coverage. However, these approaches focus on reranking/normalizing the answer probability from candidate spans after the passages have been processed independently.

\subsection{Cross-sample Information Fusion}

TaBERT \cite{tabert} proposed cross-sample attention across different rows of tabular data. Unlike TaBERT, we show that our approach is applicable to general textual data instead of vertically aligned tabular data. Cross-attention is often used in multi-modal models, to fuse information across modalities such as text and image \cite{cross_attention_1, cross_attention_2}. Most similar to our approach, \citet{vision_fusion} fuses information from multiple modalities or from different patches of the same image by fusing the CLS tokens. Our approach explores fusing information across different samples and shows that using global tokens outperforms using CLS tokens in this context.

%% file: sections/conclusion.tex
\section{Conclusion}

We propose FiE: Fusion-in-Encoder, where we extend transformer encoders with the ability to fuse information across multiple input samples using global representation via cross-sample attention, and propose a global probability space for answer spans. Facilitating this information flow across samples enables our proposed method to achieve state-of-the-art performance in 3 popular open-domain questions answering datasets by $2.5$ EM on NQ, and $0.5$ EM on TriviaQA, while simultaneously reducing parameter count and inference latency, and $4.4$ EM on WebQ. Detailed ablations and analyses further demonstrate the effectiveness of our proposed method across multiple model sizes and the capability to fuse information from a large number of samples.

% By adding ``fusion tokens'' which enable cross-sample information flow by attention over all tokens across input samples, we can calculate span probabilities in the global space of all inputs.

%% file: sections/limitations.tex
\section*{Limitations}

Adding more global representation tokens almost linearly increases our model's GPU memory usage and compute requirements. While higher number of these fusion tokens (up to $50$) results in better performance, we use a smaller number ($10$) to limit the compute and memory requirements. Furthermore, our global probability space requires classifying the embeddings for all possible spans; therefore, increasing maximum answer length will also linearly increase the number of possible spans and hence the memory, making our model unsuitable for datasets with very long answers. 

Similarly, while our model continues to improve in performance and seems to have not converged even on using 100 passages, using a higher number of passages is difficult with large models due to GPU memory constraints. Also, our extractive model is not suitable for generative QA tasks, where the model may be expected to somewhat rephrase the spans in the passages.

%% file: sections/appendix.tex
\appendix

\section{Memory, Latency and Compute Comparisons}
\label{mem_lat_comp}

Our proposed method has minimal overheads of approximately $10\%$ compared to vanilla Electra. We can verify this both empirically, as we show in \cref{tab_mem_comp_emp}, as well as theoretically, as shown in \cref{fie_overhead_eq}. The previous SOTA method R2-D2 \cite{r2d2} reported approximately $3x$ latency compared to vanilla Electra, making FiE's latency approximately $35\%$ R2-D2.

\begin{table}[h!]
\begin{center}
\begin{tabular}{lcc}
\toprule
\textbf{Model} & \textbf{TrainIter/s}  & \textbf{Mem.(GB)} \\ 
\toprule
FiD & 2.4 & 33.4 \\
Vanilla Electra & 2.6 & 35.0 \\
No Global Tokens & 2.5 & 36.1 \\
FiE & 2.3 & 37.4 \\
\bottomrule
\end{tabular}
\end{center}
\caption{Training and memory overheads of our method compared to vanilla Electra during training on NQ, bench-marked with 1 batch size (of 100 passages) on 1 A100 40GB. TrainIter/s refers to number of training iteration per second, and Mem. refers to the GPU memory utilized by the model.}
\label{tab_mem_comp_emp}
\end{table}

Let $L$ be the number of layers, $N$ be the number of input passages, each of sequence length $S$, and $G$ be the number of global fusion tokens. For a base model on NQ, these values are $L=12$, $N=100$, $S=250$, and $G=10$. The compute requirement of a vanilla transformer will approximately be :
\begin{align}
\mathrm{Vanilla} &= \mathcal{O}(LNS^2) \label{vanilla_compute_eq}
\end{align}
For FiE, because the $S$ passage tokens each attend to $S$ passage tokens and $G$ global tokens, and because the $G$ global tokens each attend to $NS$ passage tokens and $G$ global tokens, the compute will approximately be -
\begin{align}
\mathrm{FiE} &= \mathcal{O}(LNS(S+G) + LG(NS+G)) \\
&= \mathcal{O}(LNS^2 + 2LNSG + LG^2) \label{fie_compute_eq}
\end{align}

Diving \cref{fie_compute_eq} with \cref{vanilla_compute_eq}, the compute cost of FiE compared to vanilla transformer is approximately $10\%$ :
\begin{align}
\approx (1 + \frac{2G}{S}) = (1 + \frac{2*10}{250}) \approx 1.1  \label{fie_overhead_eq}
\end{align}

A contemporaneous work Zcode++ \cite{zcode_plus_plus} proposes encoder fusion by concatenating all input tokens in the last encoder layers. The last layer then has quadratic complexity of $(NS)^2$. This approach results as in a much higher overhead however, as we show below : 
\begin{align}
\mathrm{Zcode++} &= \mathcal{O}((L-1)NS^2 + (NS)^2) \\
&=\mathcal{O}(LNS^2 + (N-1)NS^2) \label{zcode_compute_eq}
\end{align}

Diving \cref{zcode_compute_eq} with \cref{vanilla_compute_eq}, the compute cost of Zcode compared to vanilla transformer is approximately $9x$ for these settings :
\begin{align}
\approx \mathcal{O}(1 + \frac{N-1}{L}) = (1 + \frac{99}{12}) \approx 9.3
\end{align}

\section{Measures of Central Tendency and Model Stability}

As our experiments are compute-intensive (due to encoding 100 context passages for every example), it is not feasible to perform multiple runs of all the experiments. We provide detailed ablations of our components instead. 

We also ran a few runs to verify the model stability to different seeds/initialization and to check the effect of training steps, the results of which we provide in \cref{tab_sup_std}. The standard error is over 2 runs for each parameter. The model's performance is not affected much by different seeds.

\begin{table}[h!]
\begin{center}
\begin{tabular}{lcc}
\toprule
\textbf{Update Steps} & \textbf{EM} & \textbf{Std Err} \\ 
\toprule
3750 & 50.9 & 0.4 \\
7500 & 50.6 & 0.2 \\
11250 & 51.7 & 0.7 \\
15000 & \textbf{52.0} & 0.2 \\
22500 & 50.6 & 0.1 \\
\bottomrule
\end{tabular}
\end{center}
\caption{Exact Match scores and Standard Error of runs with varying seeds of our method, with Electra Base model, on Natural Questions test set. Note that these runs were performed with query tokens as inputs for the global representation tokens.}
\label{tab_sup_std}
\end{table}

% \section{Effect of Higher Regularization}

\section{Hyper-Parameters}

We used the all the original model/training hyper-parameters from Electra~\cite{clark2020electra}, and data-related parameters from FiD~\cite{fid}. Hyper-parameter search was performed for the number of update steps on NQ, 2 runs for each value in \cref{tab_sup_std}. Full details of hyper-parameters are shown in \cref{tab-hyperparam}. The Context Length was restricted to 220 for the large model instead of 250 due to GPU memory constraints.

\begin{table}[h]
\begin{center}
\begin{small}
\begin{tabular}{lc}
\hline
\toprule
\textbf{Parameters} & \textbf{Values} \\
\toprule
\textit{\textbf{Optimization} }\\
Warmup steps & 10\% \\
Update Steps & 15000 [\cref{tab_sup_std}] \\
Learning rate & 5e-5(large), 1e-4(base) \\
Layerwise LR Decay & 0.9(large), 0.8(base) \\
Drop-out & 0.1 \\
Gradient clipping & 1.0 \\
Batch Size / GPU & 1 \\
Num GPUs & 8 \\
Grad Accum Steps & 8 \\
Batch Size (effective) & 64 \\
Scheduler & linear \\
\midrule
\textit{\textbf{Data/Modelling} }\\
Context Length & 220(large), 250(small/base/tiny) \\
Num Context & 100(NQ, TriviaQA), 50(WebQ)\\
Max Answer Tokens & 15 (train) \\
Max Query Tokens & 28 \\
Global Fusion Tokens & 10 \\

\bottomrule
\end{tabular}
\end{small}
\end{center}
\caption{\label{tab-hyperparam} Training Parameters.}
\end{table}

\section{Dataset Descriptions}

\begin{table}[h!]
\begin{center}
\begin{tabular}{lcccc}
\toprule
\textbf{Dataset} & \textbf{\# Train} & \textbf{\# Dev} & \textbf{\# Test} & \textbf{Recall} \\ 
\toprule
NQ & 79K & 8.8K & 3.6K & 88.7 \\
TriviaQA & 79K & 8.8K & 11K & 87.3 \\
WebQ & 3.4K & 361 & 2K & 89.5 \\
\bottomrule
\end{tabular}

\end{center}
\caption{Dataset Statistics}
\label{tab_dataset}
\end{table}

\cref{tab_dataset} provides the statistics and retrieval recall for all three datasets. The recall was calculated corresponding to top-100 documents for NQ and TriviaQA and top-50 for WebQ.

\section{WebQuestions Scores Starting from NQ}

Due to the small size of the WebQ dataset, we also initialized our model by training it on NQ. The results are reported in \cref{tab_sup_webq}. The large model surprisingly under-performs compared to the base model - we conjecture it may be because it is over-fitting on the NQ dataset. The base model score of $53.5$ on WebQ is  higher than the previous SOTA of $48.7$ by $4.8$ EM, but this is not a fair comparison as this score uses extra data.

\begin{table}[h!]
\begin{center}
\begin{tabular}{lc}
\toprule
\textbf{Model} & \textbf{WebQ}\\ 
\toprule
FiE Base & 53.5 \\
FiE Large & 52.8 \\
\bottomrule
\end{tabular}
\end{center}
\caption{EM scores on WebQ test set.}
\label{tab_sup_webq}
\end{table}

\section{Dev Scores for Corresponding Test Scores}
\cref{tab_dev_main_result}, \cref{tab_dev_ablation}, \cref{tab_dev_alternatives_fusion}, and  \cref{tab_dev_alternatives_prob} provide the dev results corresponding to the test numbers in the main paper.

\begin{table}[h!]
\begin{center}
\begin{tabular}{lccc}
\toprule
\textbf{Model} & \textbf{NQ}  & \textbf{TQA}  & \textbf{WebQ}\\ 
\toprule 
\multicolumn{4}{l}{\textit{Base Models}} \\
\textbf{FiE (Ours)} & 48.4 & 68.3  & 44.3 \\
\textbf{PAQ + FiE (Ours)}  & 50.0 & 69.5  & 51.5 \\
\midrule 
\multicolumn{4}{l}{\textit{Large Models}} \\
% \midrule 
\textbf{FiE (Ours)}  & 51.4 & 71.6 & 49.9 \\
\textbf{PAQ + FiE (Ours)}  & 53.0 & 72.7 & 50.1 \\
\bottomrule
\end{tabular}
\end{center}
\caption{EM scores on NQ, TQA and WebQ development sets for \cref{tab_main_result}.}
\label{tab_dev_main_result}
\end{table}

\begin{table}[h!]
\begin{center}
\begin{tabular}{lcc}
\toprule
\textbf{Model} & \textbf{EM}\\ 
\toprule
Electra Base (Baseline) & 45.1 \\
~~+ Global Prob Space & 47.2 \\
~~~~ + Global Repr Fusion & \textbf{48.4} \\
\bottomrule
\end{tabular}
\end{center}
\caption{Model Components Ablation on NQ with Electra Base model, EM scores on dev set for \cref{tab_ablation}.}
\label{tab_dev_ablation}
\end{table}

\begin{table}[h!]
\begin{center}
\begin{tabular}{lcc}
\toprule
\textbf{Model} & \textbf{NQ(EM)}\\ 
\toprule 
Ours & \textbf{48.5} \\
Question Repr as Fusion Tokens & 48.1 \\
All CLS attend to all CLS & 48.0 \\
Fusion Tokens only attend to CLS & 48.1 \\
No Fusion & 47.2 \\
\bottomrule
\end{tabular}
\end{center}
\caption{Exact Match scores of alternatives to our encoder fusion tokens, with Electra Base model, on Natural Questions dev set, corresponding to \cref{tab_alternatives_fusion}}
\label{tab_dev_alternatives_fusion}
\end{table}

\begin{table}[h!]
\begin{center}
\begin{tabular}{lcc}
\toprule
\textbf{Model} & \textbf{NQ(EM)}\\ 
\toprule 
Ours & \textbf{48.1} \\
Non-conditional start \& end logits & 47.8 \\
Separate start and end prob space & 46.8 \\
Span repr. sum + string prob space  & 45.3 \\
Logits sum + string prob space & 44.0 \\
\bottomrule
\end{tabular}
\end{center}
\caption{Exact Match scores of alternatives to the global probability space, with Electra Base model, on Natural Questions dev set, corresponding to \cref{tab_alternatives_prob}}
\label{tab_dev_alternatives_prob}
\end{table}

\begin{table}[h!]
\begin{center}
\begin{tabular}{lcc}
\toprule
\textbf{Model} & \textbf{NQ(EM)}\\ 
\toprule 
MML & 48.2 \\
HardEM max & 48.7 \\
HardEM min & 48.7 \\
HardEM 80\% probability mass & 48.6 \\
\bottomrule
\end{tabular}

\end{center}
\caption{Exact Match scores of alternatives to the MML training objective, with Electra Base model, on Natural Questions dev set, corresponding to \cref{tab_alternatives_obj}}
\label{tab_dev_alternatives_obj}
\end{table}

\begin{table}[h!]
\begin{center}
\begin{tabular}{llc}
\toprule
\textbf{Model} & \textbf{Params} & \textbf{NQ}\\ 
\toprule
Bert Tiny  & 4M & 26.3\\
% Bert Small
Electra Small & 14M & 42.4 \\
Electra Base & 110M & 48.4 \\
Electra Large & 330M & 51.4 \\
% Deberta Large
\bottomrule
\end{tabular}

\end{center}
\caption{Effect of Model Size on Performance (Exact Match) on NQ dev sets, corresponding to \cref{tab_abl_size}.}
\label{tab_dev_abl_size}
\end{table}

\section{Training the Bert Tiny Model}
We observed that the model trained using Bert Tiny did not converge after 15k update steps on the NQ dataset, and the dev performance was still increasing. Hence, we trained the models with different number of total steps. These results have been provided in \cref{tab_bert_tiny_steps}. The results in \cref{tab_abl_size} correspond to 50k update steps.

\begin{table}[h!]
\begin{center}
\begin{tabular}{lcc}
\toprule
\textbf{\# Steps} & \textbf{Dev} & \textbf{Test}\\ 
\toprule
15k  & 18.7 & 18.5 (+ 6.7)\\
30k & 22.1 & 21.9 (+ 7.9) \\
50k & \textbf{26.3} & \textbf{26.0} (+11.1) \\
\bottomrule
\end{tabular}
\end{center}
\caption{Bert Tiny results with different number of steps on the NQ test set. The improvements over the baselines have also been provided similar to \cref{tab_abl_size}.}
\label{tab_bert_tiny_steps}
\end{table}

\section{Raw Values for Plots}

The raw values used in the plots in \cref{tab_abl_num_glo} and \cref{tab_abl_num_pas} can be found in \cref{tab_raw_abl_num_glo} and \cref{tab_raw_abl_num_pas} respectively.

\begin{table}[h!]
\begin{center}
\begin{tabular}{lc}
\toprule
\textbf{\# Global Fusion Tokens} & \textbf{NQ} \\ 
\toprule
0 & 40.5 \\
10 & 41.6 \\
25 & 41.8 \\
50 & 42.3 \\
100 & 42.3 \\
\bottomrule
\end{tabular}
\end{center}
\caption{Effect of Number of Global Tokens on Model performance in Exact-Match on NQ dataset, with Electra-Small Model, corresponding to \cref{tab_abl_num_glo}.}
\label{tab_raw_abl_num_glo}
\end{table}

\begin{table}[h!]
\begin{center}
\begin{tabular}{lcc}
\toprule
\textbf{\# Passages} & \textbf{NQ} & \textbf{WebQ}\\ 
\toprule
1 & 33.4 & 38.2 \\
5 & 47.6 & 45.7 \\
10 & 49.1 & 47.1\\
20 & 51.3 & 48.4 \\
50 & 53.8 &  \textbf{50.8} \\
100 & \textbf{54.9} & - \\
\bottomrule
\end{tabular}
\end{center}
\caption{Effect of Number of Passages on Model performance in Exact-Match on NQ and WebQ dataset, with Electra-Base Model, corresponding to \cref{tab_abl_num_pas}.}
\label{tab_raw_abl_num_pas}
\end{table}

\section{Links to Source Code and Datasets}
\label{download_links}

The source code is based on the original implementation of FiD~\citep{fid}, which can be found at \href{https://github.com/facebookresearch/FiD/blob/25ed1ff0fe0288b80fb5e9e5de8d6346b94b8d48/README.md}{their Github}. The modeling for the fused Electra model was implemented using HuggingFace~\cite{huggingface} by modifying the \href{https://github.com/huggingface/transformers/blob/b0892fa0e8df02d683e05e625b3903209bff362d/src/transformers/modeling_electra.py}{ElectraModel}.

Data for the Wikipedia dump, Natural Questions, and TriviaQA were also downloaded from FiD's \href{https://github.com/facebookresearch/FiD/blob/25ed1ff0fe0288b80fb5e9e5de8d6346b94b8d48/README.md}{github}.

For WebQ, the QA pairs were downloaded from EMDR$^2$'s \href{https://github.com/DevSinghSachan/emdr2/blob/edb8cf6701bfa4ad7c961a21cdb461f4f7a0c72f/README.md}{github}, as well as their pre-trained WebQ checkpoint and Wikipedia context embeddings index. These were then used to retrieve the top-50 passages used in our experiments.

For PAQ, QA pairs were downloaded from PAQ's \href{https://github.com/facebookresearch/PAQ/blob/2bfd2c85e58eaac626d8d5082299a676662f51d3/README.md}{github}, and then 6M pairs were randomly selected. FiD-KD's retriever pre-trained on NQ from their \href{https://github.com/facebookresearch/FiD/blob/25ed1ff0fe0288b80fb5e9e5de8d6346b94b8d48/README.md}{github} was used to retrieve top-100 passages.

\section{Details of Evaluation Metrics}

The evaluation script from FiD was used, which can be found \href{https://github.com/facebookresearch/FiD/blob/25ed1ff0fe0288b80fb5e9e5de8d6346b94b8d48/src/evaluation.py}{here}. The evaluation metric is "Exact Match".

This metric is the average of the per-example exact match score for a dataset. The per-example exact match score is either 0 or 1; 1 if the model's answer is exactly the same as the text of any ground truth answer after lower-casing, removing punctuation, and normalizing spaces; otherwise 0.